\documentclass[11pt]{article}

\usepackage[preprint]{acl}

\usepackage{times}
\usepackage{latexsym}

\usepackage[T1]{fontenc}

\usepackage[utf8]{inputenc}

\usepackage{microtype}

\usepackage{inconsolata}

\usepackage{graphicx}

\usepackage{amssymb}
\usepackage{amsmath}
\usepackage{booktabs}

\title{RoParQ: Paraphrase-Aware Alignment of Large Language Models Towards Robustness to Paraphrased Questions}

\author{
  Minjoon Choi \\
  Seoul National University \\
  \texttt{minjoonchoi08@snu.ac.kr} \\
}

\begin{document}
\maketitle
\begin{abstract}
Large Language Models (LLMs) often exhibit inconsistent behavior when answering paraphrased questions, suggesting a reliance on surface-level patterns rather than true semantic understanding. To address this limitation, we introduce \textbf{RoParQ}, a benchmark specifically constructed to evaluate cross-paraphrase consistency in closed-book multiple-choice QA. This benchmark is derived from standard datasets by generating paraphrases via proprietary models and selectively retaining examples that elicit inconsistent confidence from a judge model. We further propose \textbf{XParaCon}, a novel evaluation metric that quantifies a model's robustness by measuring the standard deviation of accuracies across question variants. Additionally, we implement a reasoning-based, paraphrase-aware Supervised Fine-Tuning (SFT) strategy designed to align models toward semantic invariance. Our experiments demonstrate that this targeted alignment significantly enhances robustness. Notably, fine-tuned lightweight models achieved consistency levels comparable to much larger pre-trained models. These results highlight the efficacy of our approach in mitigating superficial memorization and fostering more robust, reliable LLMs.\footnote{We release our code at: https://github.com/m-joon-ixix/RoParQ}
\end{abstract}

\section{Introduction}
\label{sec:intro}

\begin{figure}[t]
    \centering
    \includegraphics[width=\columnwidth]{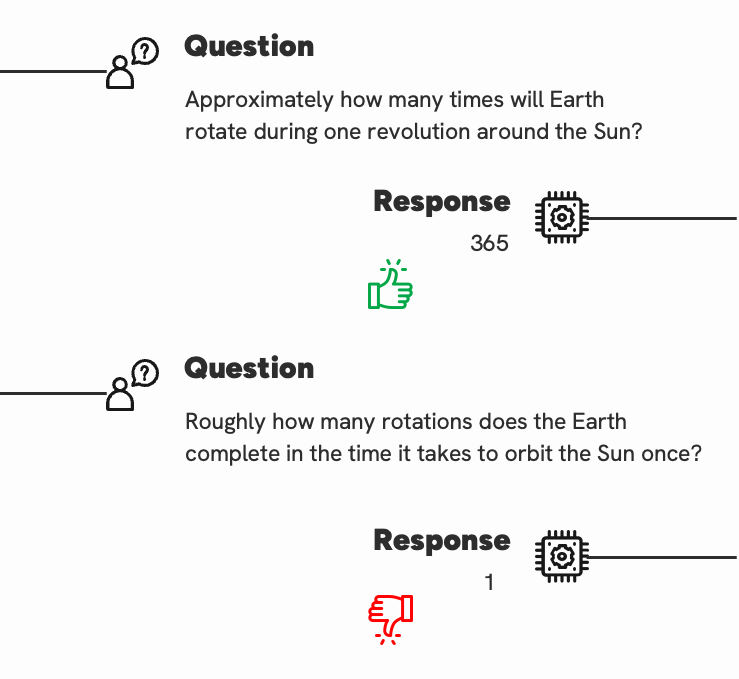}
    \caption{An example of the LLM generating an incorrect response when given a paraphrased question.}
    \label{fig:intro_figure}
\end{figure}

Large Language Models (LLMs) have demonstrated remarkable capabilities in natural language understanding and reasoning \citep{wei2022chain}, achieving high accuracy across various benchmarks \citep{brown2020languagemodels}. However, they still exhibit sensitivity to superficial variations in the given query, which is a critical limitation \citep{elazar-etal-2021-measuring, sclar2024quantifying}. Despite possessing the parametric knowledge required to answer a query correctly, LLMs often exhibit inconsistent behavior when the same question is rephrased, even when the semantic meaning remains identical, as shown in Figure \ref{fig:intro_figure}. This phenomenon suggests that their high performance on benchmarks may often stem from memorizing superficial answer patterns associated with specific phrasings \citep{mccoy2024embers}, rather than acquiring true semantic understanding \citep{berglund2024the}. This lack of robustness reveals that models may rely on surface-level cues, raising concerns about their reliability in real-world applications where user queries may be highly diverse. Addressing this inconsistency is essential for developing models that are not only accurate but also robust and semantically invariant.

In this work, we investigate and mitigate this issue by constructing a comprehensive benchmark and proposing a targeted alignment strategy. We introduce \textbf{\emph{RoParQ}} (\underline{Ro}bustness to \underline{Par}aphrased \underline{Q}uestions)\footnote{We release our benchmark through Hugging Face: https://huggingface.co/datasets/m-joon-ixix/RoParQ}, a dataset constructed by paraphrasing questions from established benchmarks (MMLU, ARC, CommonsenseQA, MathQA) using advanced proprietary models. By employing an open-source judge LLM, we specifically filtered for examples that elicited inconsistent confidence, where models answered correctly on some phrasing variants but failed to do so on others. Through this deliberate filtering, we ensure that our dataset targets the precise weakness of current LLMs. To rigorously quantify model stability, we propose a new metric, \textbf{\emph{XParaCon}} (\underline{Cross}-\underline{Para}phrase \underline{Con}sistency), which measures the standard deviation of accuracies across question variants. We first analyze the performance of various pre-trained open-source and proprietary models, observing that while robustness typically scales with parameter count, gaps in consistency across paraphrases still persist. Furthermore, we implement a reasoning-based, paraphrase-aware Supervised Fine-Tuning (SFT) approach on open-source models. Our experimental results demonstrate that this alignment technique significantly improves robustness; for instance, the Llama-3.1-8B-Instruct model's \textit{XParaCon} score rises from 2.186 to 2.629 by alignment. These results demonstrated our method’s effectiveness in enabling smaller models to achieve consistency levels comparable to, or even exceeding, those of much larger pre-trained models.

The main contributions of this paper are summarized as following:

(1) \textbf{Construction of the \emph{RoParQ} Benchmark}: We present a novel dataset enriched with high-quality paraphrases generated by proprietary models to rigorously test cross-paraphrase consistency.

(2) \textbf{Proposal of the \emph{XParaCon} Metric}: We introduce a robust evaluation metric that provides a precise and intuitive measure of a model's semantic invariance.

(3) \textbf{Efficacy of Paraphrase-Aware Alignment}: We demonstrate that our reasoning-based SFT method effectively mitigates inconsistency, showing that explicitly training models on paraphrase invariance enables lightweight models to achieve robust performance comparable to that of larger models.

\section{Task Formulation}
\label{sec:task_form}

Given the set of questions $q_i \in \mathbb{Q}$ in the dataset, we denote each question as $q_{original}$. Each $q_{original}$ is paraphrased by two proprietary models; Gemini 2.5 Flash Lite \citep{comanici2025gemini25pushingfrontier} and Claude 3.5 Sonnet \citep{Anthropic2024Claude3.5}; resulting in $q_{gemini}$ and $q_{claude}$, respectively. Eventually, a set of questions $\{q_1, q_2, q_3\}$ is formed where $q_1 = q_{original}$, $q_2 = q_{gemini}$, and $q_3 = q_{claude}$.
	
Since the task on which this study focuses is multiple-choice question answering, each data example consists of multiple choices $\mathbb{C} = \{c_1, c_2, ..., c_k\}$. To remove the effect of the ordering of choices given in the prompt, we construct 8 randomly shuffled permutations of choices: $\mathbb{C}_1, \mathbb{C}_2, ..., \mathbb{C}_8$. The prompt given to the LLM ($f$) consists of a question $q$ and a permutation of choices $\mathbb{C}_i$, so the generated response by the LLM can be denoted as $f(q, \mathbb{C}_i)$.
	
We define the LLM to be \textit{``perfectly correct''} on question $q$, if the LLM outputs the correct answer when given any of the 8 permutations of choices. This definition can be written as Equation \ref{eq:def_perfectly_correct}.
\begin{align}
    \label{eq:def_perfectly_correct}
    \forall{i} \in \{1, 2, ..., 8\}, f(q, \mathbb{C}_i) = c_{Ground Truth} \\
    \Longleftrightarrow \text{``$f$ is perfectly correct on question $q$''} \notag
\end{align}
    
Using Llama-3.1-8B-Instruct \citep{grattafiori2024llama3herdmodels} as the open-source ``judge'' LLM, we determine whether the questions $\{q_1, q_2, q_3\}$ of a certain data example are perfectly correct. If all three are perfectly correct, the LLM is considered to be ``consistently confident'' on this example. If none are perfectly correct, the case is that the model is ``consistently unconfident''. These two cases are not in the scope of interest of this study. If one or two of the questions are perfectly correct, we interpret that this example yields \textit{``inconsistent confidence''} from the LLM. Since this is the main target for investigation in our study, these examples are included in our final dataset.

\section{Experimental Setup}
\label{sec:exp_setup}

\begin{table*}[t]
  \centering
  \begin{tabular}{lccc}
    \toprule
    \textbf{} & \textbf{small-sized} & \textbf{medium-sized} & \textbf{large-sized} \\
    \midrule
    \textbf{Llama 3.1} & Llama-3.1-8B-Instruct & Llama-3.1-70B-Instruct & Llama-3.1-405B \\
    \textbf{Qwen 3} & Qwen3-4B-Instruct-2507 & Qwen3-30B-A3B-Instruct-2507 & - \\
    \textbf{Mistral} & Mistral-7B-Instruct-v0.3 & Mistral-Small-24B-Instruct-2501 & - \\
    \bottomrule
  \end{tabular}

  \caption{List of open-source LLMs used in our study.}
  \label{tab:model_list}
\end{table*}

\subsection{Datasets}
\label{sec:datasets}

The questions and multiple-choice options used to form our dataset was sourced from Unified MCQA\footnote{https://huggingface.co/datasets/pszemraj/unified-mcqa}. Among the subsets where each one represents the number of choices forming the question, we use the \textit{4-choice} and \textit{5-choice} subsets to ensure a sufficient number of options. Data examples where the source dataset is MMLU \citep{hendrycks2021measuring} or ARC \citep{clark2018thinksolvedquestionanswering} are employed from the \textit{4-choice} subset. Examples sourced from CommonsenseQA \citep{talmor-etal-2019-commonsenseqa} or MathQA \citep{amini-etal-2019-mathqa} are selected to be used from the \textit{5-choice} subset. MMLU, ARC, and CommonsenseQA are benchmarks that query for commonsense, real-world general knowledge. MathQA requires step by step reasoning involving elementary calculation in order to reach the final answer.

\subsection{Models}
\label{sec:models}

In this study, we take multiple open-source, autoregressive transformer-based LLMs into consideration. Specifically, Llama 3.1 \citep{grattafiori2024llama3herdmodels}, Qwen 3 \citep{yang2025qwen3technicalreport}, and Mistral \citep{mistral-7b-2023, mistral-small3-2025} model families are selected for analysis. In each model family, a small-sized model which has less than 10 billion parameters and a medium-sized model which has a parameter size larger than 20 billion is in the scope of our analysis, for the purpose of comparison with respect to model sizes. For the Llama 3.1 family, Llama-3.1-405B was also used in our experiments. The open-source LLMs used in our study are listed in Table \ref{tab:model_list}.
In order to compare the performance regarding robustness to paraphrased questions, we expand our experiments to Deepseek-R1 \citep{deepseekai2025deepseekr1incentivizingreasoningcapability} and two proprietary models: Gemini 2.5 Flash Lite and Claude 3.5 Sonnet\footnote{For the following models, responses were generated by accessing the inference endpoint of Snowflake Cortex API: Llama-3.1-405B, Deepseek-R1, Claude 3.5 Sonnet}.

\subsection{Evaluation}
\label{sec:evaluation}

To evaluate the correctness, or accuracy of responses generated by the LLMs, we extract the choice that was selected by the model and compare it with the ground truth label. The LLM is instructed to output its final answer after a predefined prefix\footnote{The prompts that were used to instruct LLMs are presented in Figures \ref{fig:prompt_mcqa_gk} and \ref{fig:prompt_mcqa_mr}.}, in order to eliminate the possibility of errors when parsing generated responses. The overall accuracy of responses is computed as: ${\text{(Number of correct responses)}} \div {\text{(Number of total responses)}}$.

The cross-paraphrase consistency, or robustness is measured using the standard deviation of the accuracies of the three question variants. Given each data example, we compute the accuracy for each of the three paraphrased variants of the question. By taking the standard deviation of these accuracy measures, we aim to indicate how inconsistent the LLM is in answering an identical question represented in multiple paraphrased versions. Since we obtain a standard deviation for each example in a dataset, the average of these standard deviations manifest the overall cross-paraphrase inconsistency of the whole dataset. In order to make our numeric measure truly represent the consistency itself and make the scale more intuitive, we take the log (base 2) of the average value explained above, and change its sign. As a result, we propose \textbf{\emph{XParaCon}} (\underline{Cross}-\underline{Para}phrase \underline{Con}sistency):
\begin{equation*}
    \label{eq:xparacon_definition}
    XParaCon = -log_{2}{(\frac{1}{n} \sum_{i=1}^{n} STD)}
\end{equation*}
\begin{equation*}
    \label{eq:xparacon_definition}
    STD = StdDev(acc(q_{i,0}),acc(q_{i,1}),acc(q_{i,2} ))
\end{equation*}
where $n$ is the total number of data examples, and $acc(q_{i,j})$ is the LLM’s accuracy on question $q_j$ of the $i$th data example.

\begin{table*}[t]
  \centering
  \begin{tabular}{lcccc}
    \toprule
    {} & \textbf{MMLU} & \textbf{ARC} & \textbf{CommonsenseQA} & \textbf{MathQA} \\
    \midrule
    \textbf{Initially Sourced} & 99841 & 3358 & 9741 & 29007 \\
    \textbf{After Data Preprocessing} & 3899 & 2174 & 6995 & 18542 \\
    \textbf{After Data Selection} & 707 & 344 & 2083 & 7140 \\
    \bottomrule
  \end{tabular}

  \caption{Number of data examples obtained in each step, throughout the dataset construction process.}
  \label{tab:data_size_each_step}
\end{table*}

\subsection{Reasoning-based Alignment}
\label{sec:alignment}

This study employs reasoning-based, paraphrase-aware Supervised Fine-Tuning (SFT) to align models toward semantic invariance across paraphrased questions. During SFT, each training instance includes both the original question and a paraphrased variant, and the instruction explicitly requires the model to restate the question in its own words, produce a meaning-preserving paraphrase, and verify that the same option is predicted under that paraphrase\footnote{The prompts are presented in Figures \ref{fig:prompt_para_mcqa_gk} and \ref{fig:prompt_para_mcqa_mr}.}. This design couples answer selection with an explicit reasoning routine that compares semantics across formulations, encouraging the model to ground its choice in invariant meaning rather than surface cues. By making paraphrase awareness a supervised objective, the approach leverages the model’s latent reasoning ability to reconcile multiple phrasings before committing to an answer, thereby reducing cross-paraphrase inconsistency. Through this alignment procedure that trains the model to internalize a paraphrase-consistency check as part of its decision process, we aim to improve robustness to superficial wording changes while preserving the closed-book, multiple-choice QA evaluation setting. LoRA \citep{hu2022lora} was used for Supervised Fine-tuning, and the hyperparamter settings are presented in Tables \ref{tab:sft_hyperparam} and \ref{tab:lora_hyperparam}.

\section{Dataset Construction: RoParQ}
\label{sec:dataset_const}

\subsection{Data Preprocessing}
\label{sec:data_preproc}

We enforce the closed-book multiple-choice QA task to ensure that downstream paraphrasing and robustness measurements are faithful and comparable across items. Source questions are drawn from four source datasets, as described in Section \ref{sec:datasets}. Only questions without any accompanying passages are retained to maintain a closed-book setting, eliminating items whose correctness might hinge on retrieved contexts rather than models’ parametric knowledge and inherent reasoning abilities. To remove formats incompatible with paraphrasing, all questions containing underbars are excluded since these typically indicate masked-word prediction tasks that are harder to stably paraphrase into variants with same semantic meanings. Each question must end with a question mark to reduce prompt variability, which also improves paraphrase alignment.

Questions longer than three sentences are filtered out to avoid cases where long narrative contexts are mixed in, which risks shifting the task away from closed-book QA and increases ambiguity in paraphrase generation. To ensure that paraphrasing operations are meaningful, questions shorter than ten words are removed because overly short questions tend to be underspecified and sensitive to minor rewording, thus being difficult to paraphrase without altering semantic content. After these structural filters, the retained pool consists of well-formed questions that enable controlled paraphrase generation, producing a high-integrity dataset that supports reliable measurement of cross-paraphrase stability.

\subsection{Question Paraphrasing}
\label{sec:question_para}

Proprietary models are known for their strong abilities to follow instructions, so Gemini 2.5 Flash Lite and Claude 3.5 Sonnet\footnote{Two examples from CommonsenseQA were removed since the Claude model refused to paraphrase due to harmful (porn-related) content residing in those examples.} are used for the task of question paraphrasing. For each question, we employ the following instruction to generate a paraphrased variant: \textit{``Your task is to paraphrase only the question, keeping the meaning exactly the same but using different wording''}.  Extra requirements that restrict edits only to the question and prevent any modification of choices were added to the instruction. A constraint to keep the question type (MCQA) unchanged was also augmented to avoid shifts to other tasks. The QA triplet (Question, Choices, Answer) was provided to the LLM in order to ground the paraphrase in the precise decision context. The exact prompt that was used to query LLMs to paraphrase questions is provided in Figure \ref{fig:prompt_question_paraphrase}.

\begin{table*}[t]
    \centering
    \begin{tabular}{lcccc}
        \toprule
        {} & \multicolumn{2}{c}{General Knowledge} & \multicolumn{2}{c}{Math Reasoning} \\
        \cmidrule(lr){2-3} \cmidrule(lr){4-5}
        \textbf{Model} & \textbf{Accuracy} & \textbf{XParaCon} & \textbf{Accuracy} & \textbf{XParaCon} \\
        \midrule
        Llama-3.1-8B-Instruct & 0.781 & 2.186 & 0.738 & 1.924 \\
        Llama-3.1-70B-Instruct & 0.869 & 3.219 & 0.863 & 3.114 \\
        Llama-3.1-405B & \textbf{0.882} & 3.272 & 0.934 & 3.762 \\
        Mistral-7B-Instruct-v0.3 & 0.697 & 2.663 & 0.344 & 2.728 \\
        Mistral-Small-24B-Instruct-2501 & 0.855 & 3.121 & 0.899 & 3.435 \\
        Qwen3-4B-Instruct-2507 & 0.802 & 2.848 & \underline{0.942} & 4.489 \\
        Qwen3-30B-A3B-Instruct-2507 & 0.866 & 3.136 & 0.935 & 4.034 \\
        Deepseek-R1 & 0.850 & \underline{3.378} & 0.915 & 4.405 \\
        Claude 3.5 Sonnet & \underline{0.876} & \textbf{3.428} & \textbf{0.959} & \textbf{5.164} \\
        Gemini 2.5 Flash Lite & 0.861 & 3.339 & 0.941 & \underline{4.195} \\
        \bottomrule
    \end{tabular}
    \caption{Performance of pre-trained open-source models and proprietary models on each subset of our benchmark RoParQ. The highest value in each column is highlighted in \textbf{bold}, and the second-highest value is \underline{underlined}.}
    \label{tab:perf_base}
\end{table*}

\subsection{Data Selection}
\label{sec:data_select}

To execute the data selection process, we generate responses for each question variant and random-shuffled permutation of choices. As a result, we generate\footnote{Greedy decoding (temperature = 0) was used to generate responses.} 24 responses for each data example, using our judge model Llama-3.1-8B-Instruct. Meanwhile, the consistent accuracy statistics of these generated responses\footnote{Accuracy statistics are presented in Table \ref{tab:data_sel_performance}.} across $q_{original}$, $q_{gemini}$, and $q_{claude}$ depict the above par quality of the question paraphrasings. As described in Section \ref{sec:task_form}, examples in which the judge LLM was \textit{``perfectly correct''} on one or two question variants, which is defined as examples yielding \textit{``inconsistent confidence''}, are added to our final dataset, \textbf{\emph{RoParQ}} (\underline{Ro}bustness to \underline{Par}aphrased \underline{Q}uestions). The number of examples obtained in each step; initially sourced, after data preprocessing, and after data selection; is shown in Table \ref{tab:data_size_each_step}.

Since each source dataset exhibits different characteristics (Section \ref{sec:datasets}), we categorize RoParQ into two subsets: \emph{general knowledge} and \emph{math reasoning}. Examples with questions asking for general knowledge or commonsense (MMLU, ARC, and CommonsenseQA) reside in the \emph{general knowledge} subset. On the other hand, examples with queries that require deeper, multi-step numerical reasoning (MathQA) are labeled as the \emph{math reasoning} subset.	Furthermore, our benchmark RoParQ consists of three splits: train (70\%), validation (15\%), and test (15\%). The number of examples in each subset and split is presented in Table \ref{tab:data_size_subset_split}.

\begin{table}[h]
  \centering
  \small
  \begin{tabular}{lccc}
    \toprule
    {} & \textbf{Train} & \textbf{Validation} & \textbf{Test} \\
    \midrule
    \textbf{General Knowledge} & 2194 & 470 & 470 \\
    \textbf{Math Reasoning} & 4998 & 1071 & 1071 \\
    \bottomrule
  \end{tabular}

  \caption{Number of data examples in each subset/split setting of the RoParQ benchmark.}
  \label{tab:data_size_subset_split}
\end{table}

\section{Results}
\label{sec:results}

\subsection{Baselines}
\label{sec:results_base}

We first present the overall accuracy of responses and \textit{XParaCon} metrics on pre-trained open-source models and proprietary models, in Table \ref{tab:perf_base}.

The performance analysis across the two RoParQ subsets revealed distinct patterns. On the \emph{general knowledge} subset, the largest open-source model, Llama-3.1-405B, achieved the highest accuracy, followed by Claude 3.5 Sonnet. However, Claude 3.5 Sonnet demonstrated the strongest cross-paraphrase consistency, closely followed by Gemini 2.5 Flash Lite, suggesting superior robustness among the proprietary models in this domain. The trend of scaling performance is especially evident within the Llama 3.1 family, where the 70B model significantly outperforms the 8B model in both accuracy and \textit{XParaCon}, and the 405B model further improves upon both metrics, reinforcing the general principle that performance scales with model size.

\begin{table*}[t]
    \centering
    \begin{tabular}{lcccc}
        \toprule
        {} & \multicolumn{2}{c}{General Knowledge} & \multicolumn{2}{c}{Math Reasoning} \\
        \cmidrule(lr){2-3} \cmidrule(lr){4-5}
        \textbf{Model} & \textbf{Accuracy} & \textbf{XParaCon} & \textbf{Accuracy} & \textbf{XParaCon} \\
        \midrule
        Llama-3.1-8B-Instruct & 0.781 & 2.186 & 0.738 & 1.924 \\
        Llama-3.1-8B-Instruct \textbf{(FT)} & 0.798 & \textbf{2.629} & 0.685 & \textbf{2.316} \\
        Mistral-7B-Instruct-v0.3 & 0.697 & 2.663 & 0.344 & 2.728 \\
        Mistral-7B-Instruct-v0.3 \textbf{(FT)} & 0.735 & \textbf{2.854} & 0.405 & \textbf{2.617} \\
        Qwen3-4B-Instruct-2507 & 0.802 & 2.848 & 0.942 & 4.489 \\
        Qwen3-4B-Instruct-2507 \textbf{(FT)} & 0.821 & \textbf{2.920} & 0.951 & \textbf{4.856} \\
        \bottomrule
    \end{tabular}
    
    \caption{Performance of open-source models (pre-trained and fine-tuned) on each subset of our benchmark RoParQ. Fine-tuned models are denoted as `\textbf{FT}'. \textit{XParaCon} scores of fine-tuned models are highlighted in \textbf{bold}.}
    \label{tab:perf_finetuned}
\end{table*}

\begin{figure*}[t]
    \centering
    \includegraphics[width=\textwidth]{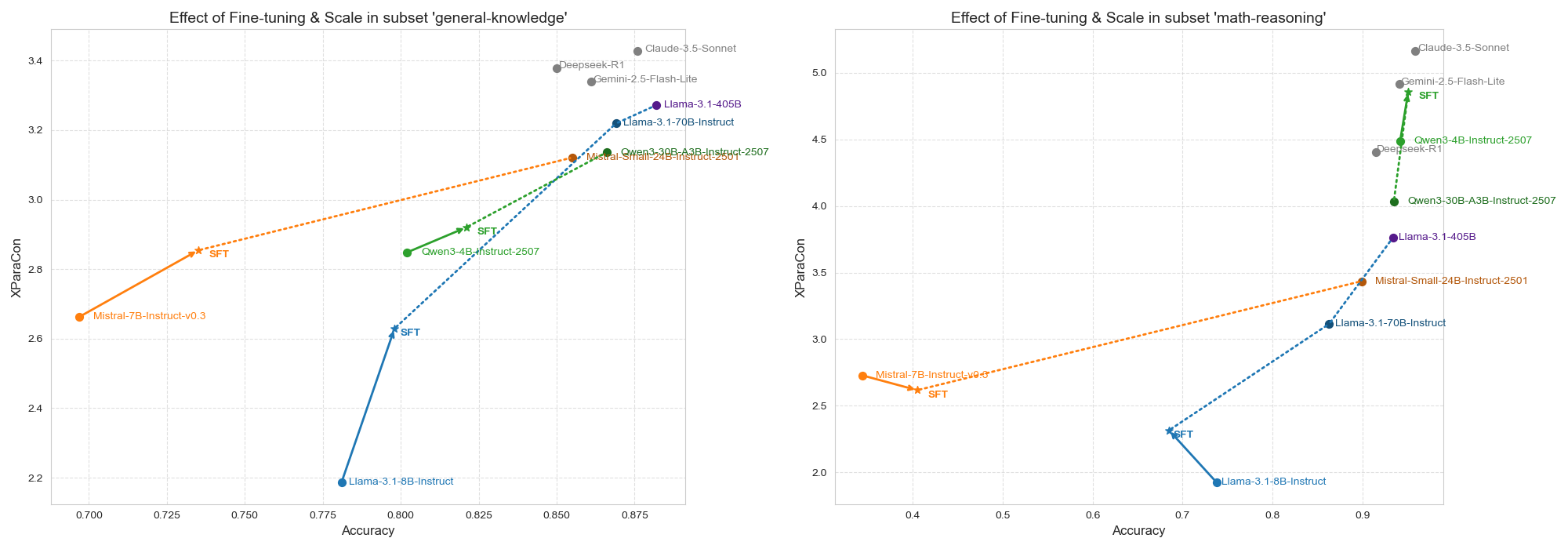}
    \caption{Movements of Accuracies and \textit{XParaCon} scores within each model family, representing the effect of scale and fine-tuning.}
    \label{fig:2d_scatter_plot}
\end{figure*}

Conversely, for the \emph{math reasoning} subset, Claude 3.5 Sonnet was the top performer in both metrics, achieving the highest accuracy and a considerably higher \textit{XParaCon}, with Gemini 2.5 Flash Lite holding the second rank for robustness. These results collectively indicate that the proprietary models, particularly Claude 3.5 Sonnet, generally offer the most consistent and robust performance against paraphrasing, especially in challenging math reasoning tasks, although the largest open-source models remain highly competitive in general knowledge accuracy. A notable exception to the overall hierarchy is the Qwen3-4B-Instruct-2507 model. Despite its relatively small parameter count, this model achieved the second-highest accuracy among all models and a strong \textit{XParaCon} score, suggesting that specialized training techniques, such as strong-to-weak distillation used in the Qwen family, can lead to unexpectedly efficient and effective reasoning capabilities in resource-constrained models for specific tasks.

\subsection{Post-Alignment}
\label{sec:results_align}

\begin{figure*}[t]
    \centering
    \includegraphics[width=\textwidth]{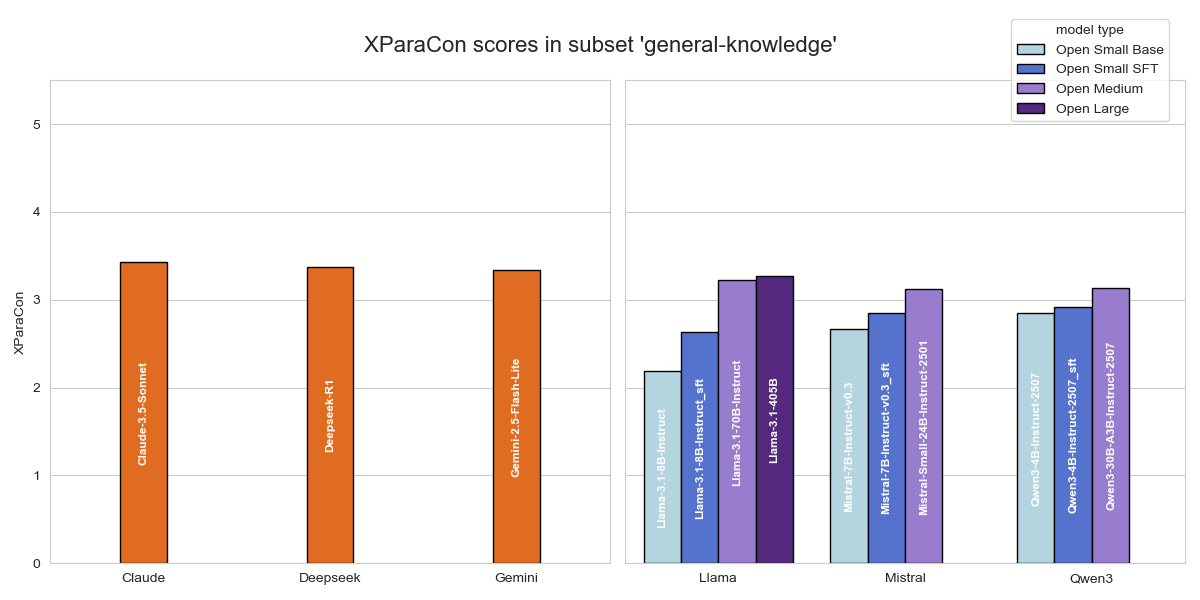}
    \caption{\textit{XParaCon} score of each model in the \textit{general knowledge} subset.}
    \label{fig:xparacon_barplot_gk}

    \vspace{0.5cm}

    \includegraphics[width=\textwidth]{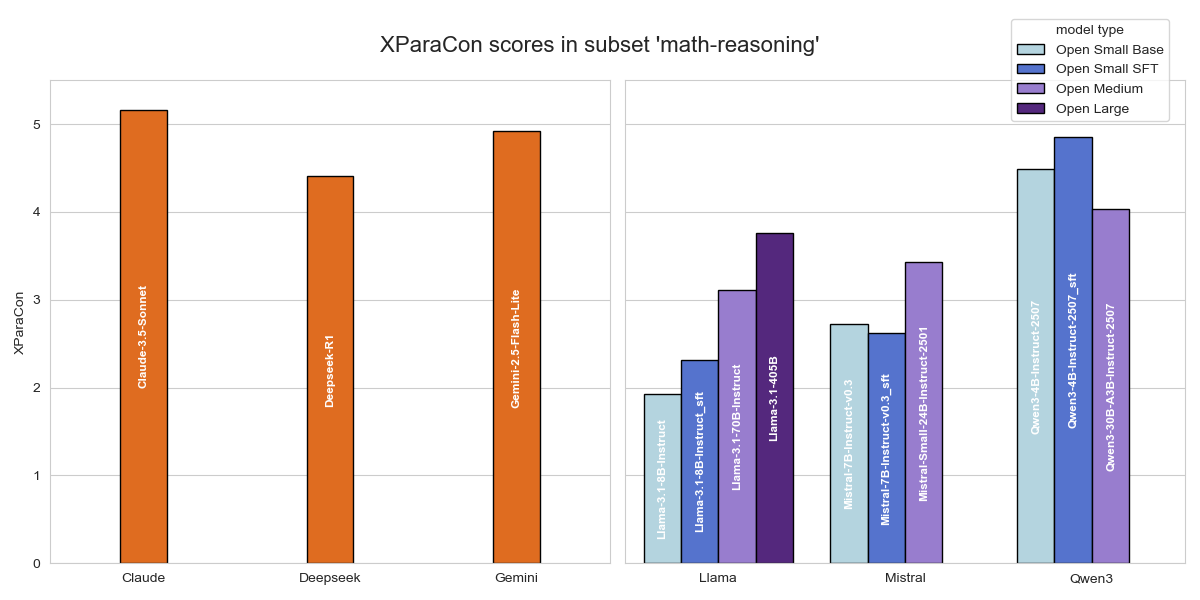}
    \caption{\textit{XParaCon} score of each model in the \textit{math reasoning} subset.}
    \label{fig:xparacon_barplot_mr}    
\end{figure*}

Table \ref{tab:perf_finetuned} describes the overall accuracy and \textit{XParaCon} of small-sized open-source models and their fine-tuned checkpoints.\footnote{Figures \ref{fig:xparacon_barplot_gk} and \ref{fig:xparacon_barplot_mr} provide a comprehensive view of the \textit{XParaCon} scores for each model.}

The results across both subsets confirm the efficacy of our fine-tuning based approach in improving model robustness to paraphrased questions. On the \emph{general knowledge} subset, fine-tuning yielded a consistent and substantial increase in \textit{XParaCon} scores for all models (Figure \ref{fig:2d_scatter_plot}), often bringing the smaller fine-tuned models’ scores near those of much larger pre-trained models in the same family, effectively leveraging targeted training to compensate for a smaller parameter count. A similar trend was observed in the \emph{math reasoning} subset, where the Qwen3-4B model, for example, saw its \textit{XParaCon} score improve from 4.489 to 4.856. This fine-tuned robustness, particularly for Qwen3-4B, reached a level comparable to, or exceeding, several much larger open-source models, underscoring the power of goal-oriented supervised fine-tuning. However, it is noteworthy that for the Llama-3.1-8B-Instruct model on the \emph{math reasoning} subset, this significant boost in \textit{XParaCon} was accompanied by a minor decline in raw accuracy, lighting the possibility of a potential trade-off between consistency and correctness in domain-specific tasks for light-weighted models.

\section{Related Work}
\label{sec:rel_work}

\citet{geusau-2021-evaluating} evaluate QA robustness to paraphrased questions, finding consistent performance drops and showing BERT’s sentence embeddings better capture paraphrase similarity than alternatives. Unlike their focus on extractive QA and broad paraphrase coverage, our work targets closed-book multiple-choice QA and focuses on data examples exhibiting cross-paraphrase inconsistency, enabling fine-grained analysis. \citet{gan-ng-2019-improving} create two SQuAD-based paraphrase datasets and show significant degradation across state-of-the-art models. They also explore data augmentation via neural paraphrasing to improve robustness. In contrast, this work selects only questions that empirically trigger inconsistency across paraphrases, and further proposes a paraphrase-aware supervised fine-tuning protocol aimed at enforcing answer invariance. \citet{deng2024rephraserespondletlarge} propose Rephrase-and-Respond (RaR), where one LLM rephrases prompts and another generates the answer, improving reliability across diverse tasks. While RaR uses self-rephrasing as an inference-time strategy to enhance performance, our work builds a benchmark to evaluate robustness to paraphrases and introduces a training-time method to align model behavior across paraphrases. \citet{dong-etal-2017-learning} present a framework that generates candidate question paraphrases from multiple sources and learns a neural scoring model to weight paraphrases in order to improve QA performance. In contrast, our work proposes a paraphrase-aware supervised fine-tuning protocol to enforce answer invariance across question variants, rather than learning to score paraphrases within a retrieval or span-extraction pipeline.

\section{Conclusion}
\label{sec:conclusion}

This study was motivated by the  limitations of Large Language Models regarding their sensitivity to superficial input variations. To address these limitations, we constructed the \textbf{\emph{RoParQ}} benchmark to rigorously test cross-paraphrase consistency and introduced the \textbf{\emph{XParaCon}} metric to quantify it. Furthermore, we proposed a reasoning-based Supervised Fine-Tuning strategy aimed at aligning models toward semantic invariance. Our analysis initially confirmed that robustness generally scales with parameter count. Moreover, our experiments validate the efficacy of our alignment approach, showing that fine-tuning significantly boosts consistency scores. Notably, a fine-tuned model achieved robustness  comparable to that of much larger baselines, demonstrating that targeted alignment can effectively liberate semantic stability from the constraints of parameter scale. This establishes a promising direction for developing efficient, reliable LLMs that do not rely on massive scale alone for robustness. Moving forward, this work serves as a foundation for further research into paraphrase-aware training, ultimately steering the field toward models that possess a genuine, generalized understanding of language beyond surface-level patterns.

\section*{Limitations}
\label{sec:limitations}

Our study has limitations that guide future research. Firstly, we focused primarily on commonly accessible LLM parameter sizes, reserving the evaluation of significantly larger open-source models for future work, which is necessary to fully assess the potential impact of scalability on robustness. Secondly, while effective, the RoParQ benchmark is limited to multiple-choice questions. Extending it to open-ended questions would broaden its scope. Furthermore, the evaluation was conducted solely in English, necessitating future work to explore robustness in a multilingual context. Our study relied on Supervised Fine-Tuning (SFT). While SFT was successful at dramatically improving consistency scores, future investigations may explore more advanced post-training methods, such as Reinforcement Learning from Human Feedback (RLHF) or other alignment techniques.

\section*{Acknowledgments}

This work was supported by the Korea Institute for Advancement of Technology (KIAT) grant funded by the Korean Government (Ministry of Education). (P0025681-G02P22450002201-10054408, Semiconductor-Specialized University)

\bibliography{custom}

\appendix

\section{Appendix}
\label{sec:appendix}

\begin{figure*}[h]
    \centering
    \includegraphics[width=\textwidth]{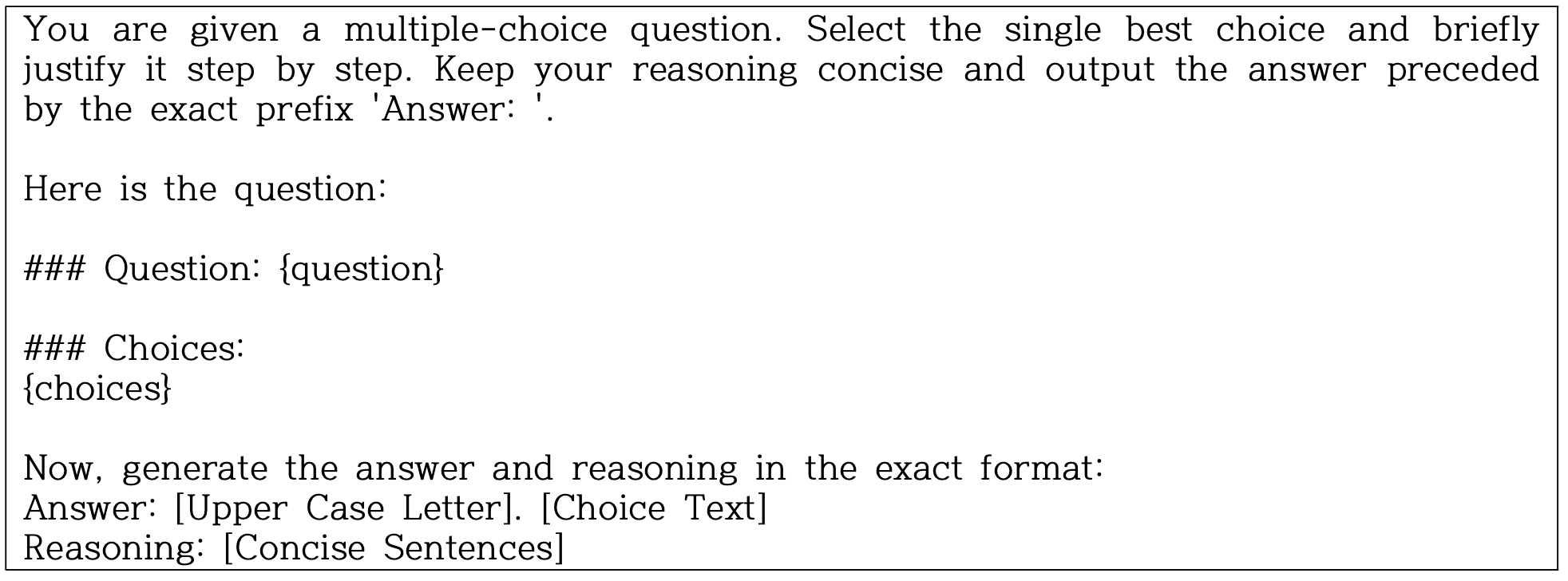}
    \caption{Prompt used for multiple choice question answering in the \textit{general knowledge} subset.}
    \label{fig:prompt_mcqa_gk}
\end{figure*}

\begin{figure*}[h]
    \centering
    \includegraphics[width=\textwidth]{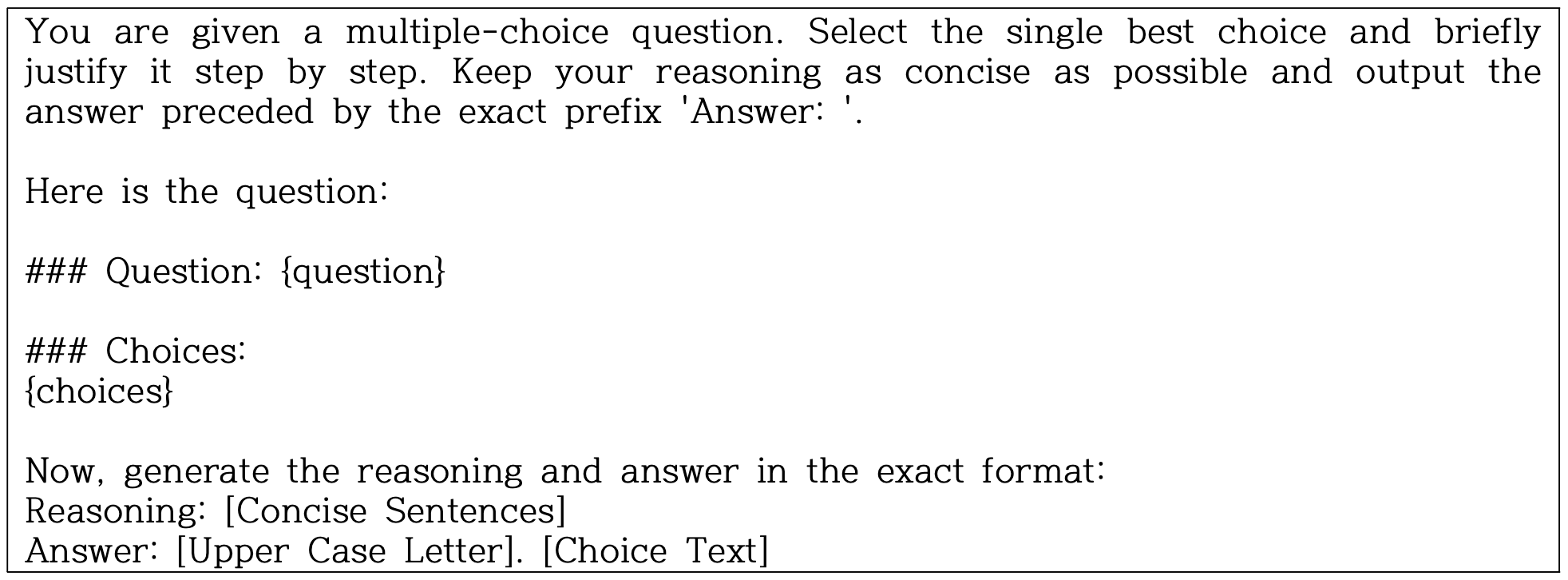}
    \caption{Prompt used for multiple choice question answering in the \textit{math reasoning} subset. The model is instructed to generated its reasoning first since questions in this subset necessarily require step by step reasoning.}
    \label{fig:prompt_mcqa_mr}
\end{figure*}

\begin{figure*}[h]
    \centering
    \includegraphics[width=\textwidth]{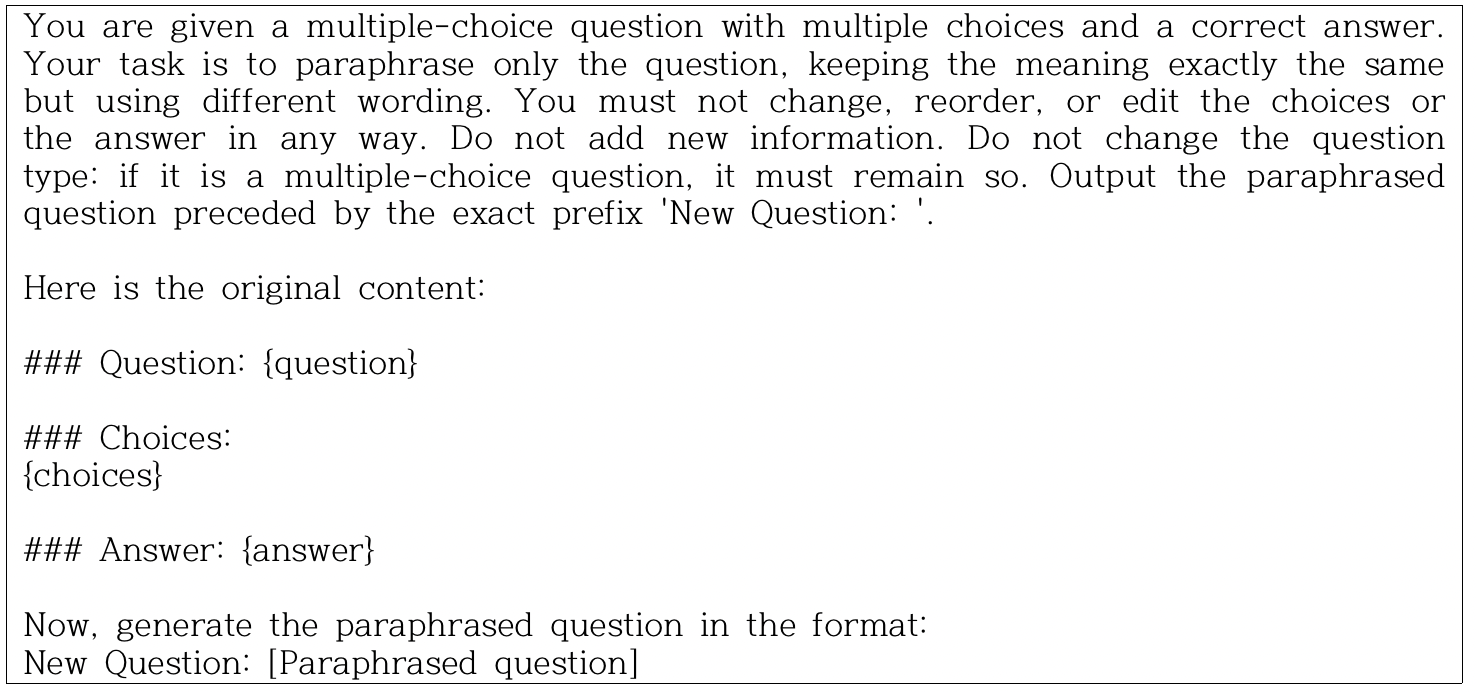}
    \caption{Prompt used for question paraphrasing. The model is instructed to output
the paraphrased question in a fixed format.}
    \label{fig:prompt_question_paraphrase}
\end{figure*}

\begin{figure*}[h]
    \centering
    \includegraphics[width=\textwidth]{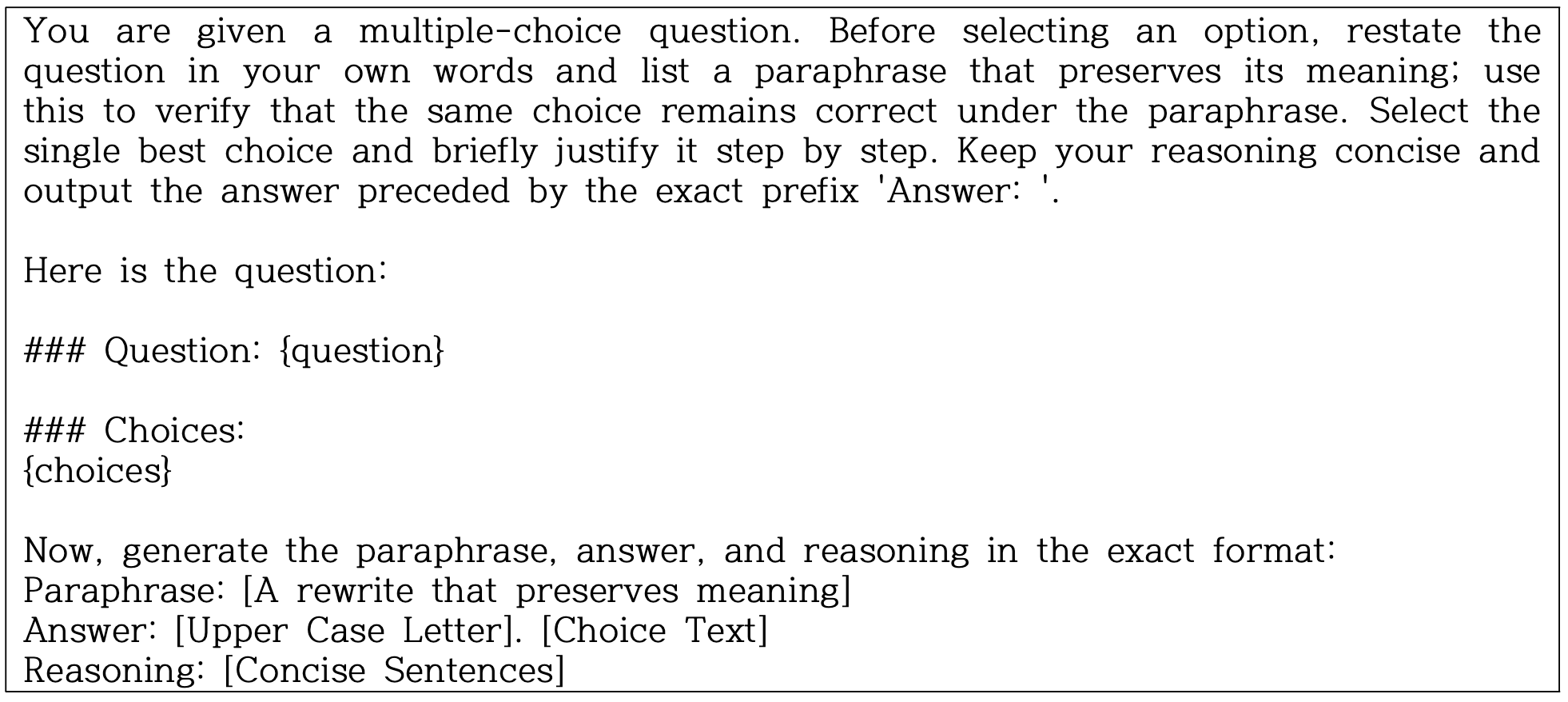}
    \caption{Prompt used for paraphrase-aware multiple choice question answering in the \textit{general knowledge} subset. This prompt was used for fine-tuning and inference on fine-tuned models.}
    \label{fig:prompt_para_mcqa_gk}
\end{figure*}

\begin{figure*}[h]
    \centering
    \includegraphics[width=\textwidth]{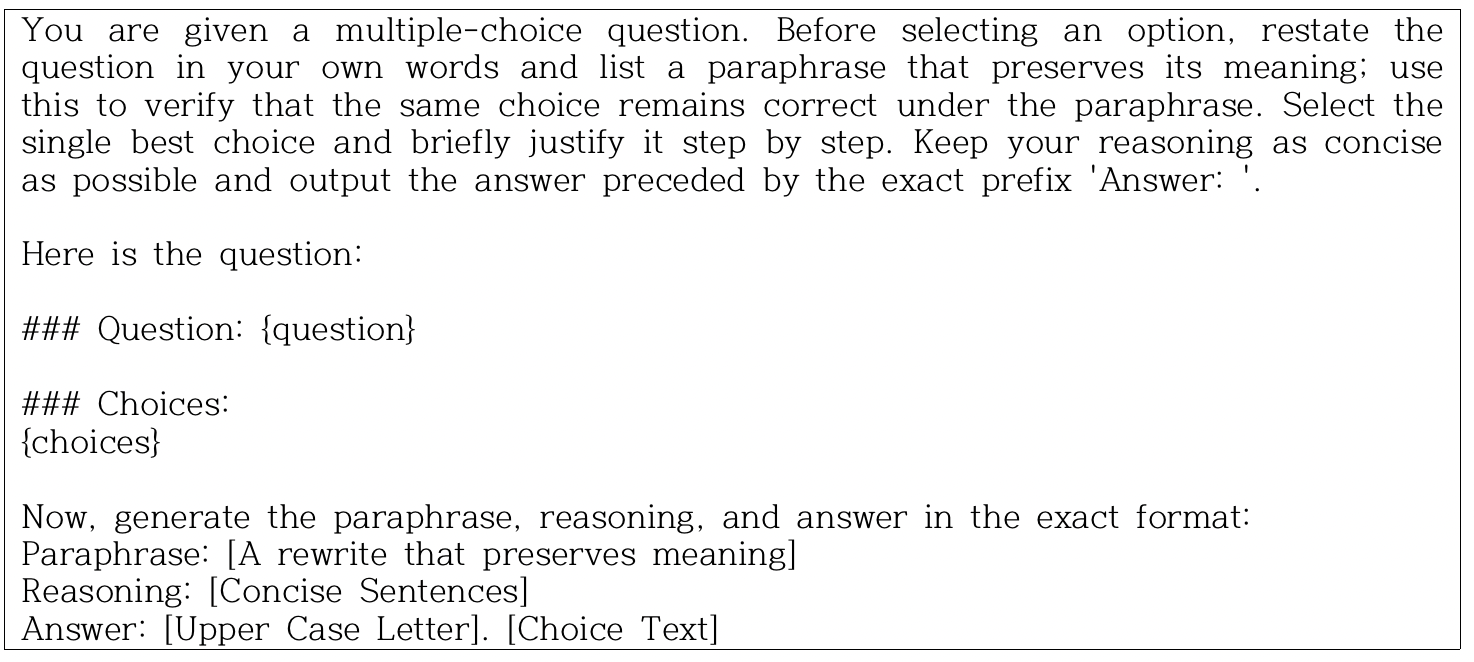}
    \caption{Prompt used for paraphrase-aware multiple choice question answering in the \textit{math reasoning} subset. This prompt was used for fine-tuning and inference on fine-tuned models.}
    \label{fig:prompt_para_mcqa_mr}
\end{figure*}

\begin{table}[h]
    \centering
    \small
    \begin{tabular}{lc}
    \toprule
    \textbf{Hyperparameter} & \textbf{Value} \\
    \midrule
    {seed} & {42} \\
    {learning\_rate} & {0.0002} \\
    {lr\_scheduler\_type} & {linear} \\
    {per\_device\_train\_batch\_size} & {1} \\
    {per\_device\_eval\_batch\_size} & {1} \\
    {warmup\_ratio} & {0.03} \\
    {weight\_decay} & {0} \\
    {logging\_strategy} & {steps} \\
    {logging\_steps} & {100} \\
    {eval\_strategy} & {steps} \\
    {eval\_steps} & {500} \\
    {save\_strategy} & {steps} \\
    {save\_steps} & {500} \\
    {save\_total\_limit} & {2} \\
    {load\_best\_model\_at\_end} & {TRUE} \\
    {metric\_for\_best\_model} & {eval\_loss} \\
    {greater\_is\_better} & {FALSE} \\
    {gradient\_accumulation\_steps} & {1} \\
    {max\_grad\_norm} & {1} \\
    {bf16} & {TRUE} \\
    \bottomrule
    \end{tabular}
  
    \caption{List of hyperparameters used for Supervised Fine-tuning (SFT).}
    \label{tab:sft_hyperparam}
\end{table}

\begin{table}[h]
    \centering
    \small
    \begin{tabular}{lc}
    \toprule
    \textbf{Hyperparameter} & \textbf{Value} \\
    \midrule
    r & 16 \\
    lora\_alpha & 32 \\
    lora\_dropout & 0.05 \\
    bias & none \\
    task\_type & CAUSAL\_LM \\
    \bottomrule
    \end{tabular}
    
    \caption{List of LoRA hyperparameters used for Supervised Fine-tuning (SFT).}
    \label{tab:lora_hyperparam}
\end{table}

\begin{table*}[h]
    \centering
    \begin{tabular}{lcccccc}
    \toprule
    {} & \multicolumn{3}{c}{\textbf{General Knowledge}} & \multicolumn{3}{c}{\textbf{Math Reasoning}} \\
    \cmidrule(lr){2-4} \cmidrule(lr){5-7}
    {} & $q_{original}$ & $q_{gemini}$ & $q_{claude}$ & $q_{original}$ & $q_{gemini}$ & $q_{claude}$ \\
    \midrule
    \textbf{Overall Accuracy} & 0.785 & 0.770 & 0.792 & 0.597 & 0.538 & 0.582 \\
    \textbf{Ratio of Perfectly Correct Examples} & 0.646 & 0.636 & 0.664 & 0.380 & 0.317 & 0.369 \\
    \bottomrule
    \end{tabular}
    
    \caption{Accuracy statistics on responses generated by Llama-3.1-8B-Instruct to select examples that yield inconsistent confidence.}
    \label{tab:data_sel_performance}
\end{table*}

\end{document}